# Comparative analysis of machine learning and numerical modeling for combined heat transfer in Polymethylmethacrylate

Mahsa Dehghan Manshadi, Nima Alafchi, Alireza Taat, Milad Mousavi, Amir Mosavi*

Faculty of Engineering, Obuda University, Budapest, Hungary

**Abstract:** This study compares different methods to predict the simultaneous effects of conductive and radiative heat transfer in a Polymethylmethacrylate (PMMA) sample. PMMA is a kind of polymer utilized in various sensors and actuator devices. One-dimensional combined heat transfer is considered in numerical analysis. Computer implementation was obtained for the numerical solution of governing equation with the implicit finite difference method in the case of discretization. Kirchhoff transformation was used to get data from a non-linear equation of conductive heat transfer by considering monochromatic radiation intensity and temperature conditions applied to the PMMA sample boundaries. For Deep Neural Network (DNN) method, the novel Long Short-Term Memory (LSTM) method was introduced to find accurate results in the least processing time than the numerical method. A recent study derived the combined heat transfers and their temperature profiles for the PMMA sample. Furthermore, the transient temperature profile is validated by another study. A comparison proves a perfect agreement. It shows the temperature gradient in the primary positions that makes a spectral amount of conductive heat transfer from a PMMA sample. It is more straightforward when they are compared with the novel DNN method. Results demonstrate that this artificial intelligence method is accurate and fast in predicting problems. By analyzing the results from the numerical solution, it can be understood that the conductive and radiative heat flux is similar in the case of gradient behavior, but it is also twice in its amount approximately. Hence, total heat flux has a constant value in an approximated steady-state condition. In addition to analyzing their composition, ROC curve and confusion matrix were implemented to evaluate the algorithm's performance.

## 1. Introduction

In recent decades, communication in human beings has increased. Because of the high growth rate, this issue needs to be improved by communicating systems and data transferring. The most suitable material for these applications is optical fiber, so the researchers have studied the different types of composite optical fibers, especially plastic optical fiber (POF) [1,2]. Beyond these activities, the transmission medium's production was stopped for a few years because of the lack of commercial field claims. Composite fiber has various applications such as light transmission sensing, data transmissions for optical signals, and deformation detection, which is used in solid mechanical experimental researches and performs chemical sensing [3-6]. Their applications are critical because they have improved the reliability of electrical machines, ICs, and microelectronics. Nevertheless, this research focused on using the polymethylmethacrylate (PMMA) plastic optical fiber (PMMA-POF) for data transferring.

POF is one of the polymer categories, which can easily be integrated into the textile statement. One of the mechanical properties of POF is having electromagnetic radiation and does not have any heat generation [7-9]. Related to the above, researchers focus on the penalties that POF may have on the performance of the transmissions. The physical characteristics of PMMA-POF are the core feature size of 1.49 and 1.59 millimeters. Utilizing silicone resin was investigated with an approximate refractive index of 1.46 millimeters for fiber cladding, maintaining a high difference between core and cladding, and using mechanical flexibility. Attenuation loss is about 1 nm at 650 nm, and the total price is decreased [8].

One of the crucial factors for determining the maximum length of the fiber link is attenuation. It depends on a wavelength and is also related to its material properties. Three types of windows can be named here related to the attenuation values. They exist around 500 nm, 570 nm, and 650 nm, starting from at least 80 dB/km. If the value ranges are in the visible wavelength, colors will show these windows, predominantly blue-green, yellow, and red. The low attenuation is related to the green and blue windows even though there are lower values related to the yellow color, which is neglected here because of low magnitudes; still, the red one has a higher value at a higher rate of speed [9]. Furthermore, it can ultimately affect heat transfer. Conductive heat transfer is extensively impressed by the physical properties of a material. However, the selected material for this study is the optimum one in the case of combined heat transfer.

For the PMMA-POF, the length of transmission becomes limited to the range of ten or a hundred meters which depends on the baud rate. For the diverse applications of data communications over PMMA-POF, the milestone market is introduced as the home or office networks. When the system is going to dimension and then installed, the bending loss becomes an essential parameter. For 360° bending, the different values of extra-losses are determined by the equilibrium model. It can be considered that the bending extra-attenuation is near 0.5 dB for a 10 mm bending radius of PMMA-POF material, while the extra-loss is near zero when the bending radius becomes at least 25mm [9]. By selecting the PMMA-POF material, energy can easily be transferred by combined radiation and conduction heat transfer internally because of the high data transfer speed. Consequently, radiation role as the heat transfer type is more important than the conductive heat transfer[9].

Multitudes of researches have done the combined heat transfer effect in the glass. Hemath [10] studied the fiber conductivity of reinforced composites by considering radiation. Sallam [11] studied a parameter that can affect radiation shielding parameters. Li et al. [12] studied thermal performance evaluation of glass windows combining silica aerogels and phase change materials for the cold climate of China. Barnoss et al. [13] analyzed transient heat transfer in a semitransparent area to predict the temperature distribution, and also they predicted conduction and radiation transient heat transfer in high temperatures [14]. Malek et al. [15] studied a transient numerical thermal radiation solution on a media. Numerical solution of Rosseland model for transient thermal radiation in non-grey optically thick media using enriched basis functions. Prerana and Satapathy [16,17] investigated the transient conduction and radiation problem using the finite volume method. Wakif A. analyzed the convection heat transfer on steady flow by considering the radiative effect on a horizontal sheet. O.D. Makinde [18] analyzed the mass transfer past a moving vertical porous plate with a thermal radiation effect. Also, Chu et al. [19] studied the combined impacts of the thermal dependence in convection heat transfer by considering radiative heat flux boundary conditions.

Among all of the tools used in the industry, machine learning with artificial neural network (ANN) and internet of things (IoT) is a powerful modeling and forecasting tool that offers an alternative way to solve complex problems such as predicting production capacity [20]. Also, several studies were conducted in the case of different machine learning methods approximations in a heat transfer field [21-23]. Ju and Shiomi [24] studied the material informatic in a case of heat transfer applications. They discuss recent progress in developing materials informatics (MI) for heat transport. They became successful in order to discuss recent progress in developing materials informatics. Their attempt presents that these methods are the most beneficial for designing thermal functional materials. Also, Ju et al. [25] also studied the coupled machine learning and thermal transport effects for designing a material with a thermal function. They claim a gap between big data and few data collected from numerical or experimental modeling. So, transfer learning here plays a vital role in filling this gap. The utilized machine learning algorithm was Bayesian optimization algorithms. Furthermore, they have introduced future fields of study.

In this study, an approximation for evaluating the radiative and conductive heat transfer in a PMMA sample is developed. A finite difference scheme is used to obtain a model of the combined heat transfer. The equations are derived for a glass optical fiber, PMMA subjected to different boundaries. The temperature distribution and the fluxes two named heat transfer types heat are obtained. A comparison is made with the solution of computer implementation and the data was used in an LSTM algorithm to find the best correlation between all of the relative heat transfer parameters.

## 2. Materials and Methods

*2.1 Numerical analysis*

This research focused on the conductivity of PMMA-POF measurements at different temperatures. In general, GOF consists of three main parts called layers. The inner layer is named "core transfers data." The next layer covers the core and should limit the light transmission in a line. These two layers are made of pure silica glass. The outer layer has been named a coating that protects the inner ones and makes of plastic or acrylate covering. However, for the POF, the core material is general-purpose resin, and the core-covering (clad) is made of fluorinated polymer [26]. Also, for PMMA-POF, two wires made of tantalum and 25 μm diameters are used here as experimental modeling.

For measuring the thermal conductivity, electricity is used. When there is a current in the wires for a short period (5s), they will get heat by their transient temperature. That means when the PMMA-POF becomes hot, a significant internal emission will happen. It means that energy is transferred from one layer to another by the radiation effect. Thus, the transient response of the thermal effect is much different when it is combined with heat transfer [27]. For measuring the resistance of a wire, the computer program of the Wheatstone bridge is used [28].

The two equations of conductive heat transfer and radiative one should simultaneously solve the case of combined heat transfer. As mentioned before, the sample's material PMMA-POF has time-dependent physical properties, anisotropic absorbed, non-gray material, emission, and scatter [29]. The assumption for the front face is a shock temperature, and the back face is a uniform temperature. Also, the medium boundaries temperature is assumed, and the initial temperature is assumed to be uniform.

For radiation analysis of PMMA-POF wire, several considerations should be applied to theoretical modeling. Although many methods exist for analyzing the radiation heat transfer, the monochromatic radiation intensity $J\_(\lambda (X.\mu.t))$ is obtained from the radiative heat transfer equation (RTE) [29]. Eq.1 is written for the position x in a μ direction with a λ wavelength at time t.

$$\frac{1}{c}\frac{\partial I_\lambda(X,\mu,t)}{\partial t} + \mu\frac{\partial I_\lambda(X,\mu,t)}{\partial X} = +J_\lambda(X,\mu,t) - \beta_\lambda(\mu)I_\lambda(X,\mu,t) \quad (1)$$

By using the order of magnitude, the first term in Eq.1 can be neglected because the propagation speed c is higher than the other terms. So the Eq.2 can be written as follows;

$$\mu\frac{\partial I_\lambda(X,\mu,t)}{\partial X} = -\beta_\lambda(\mu)I_\lambda(X,\mu,t) + J_\lambda(X,\mu,t) \quad (2)$$

The source function defined as an increasing $J_\lambda(X,\mu,t)$ radiation intensity per unit thickness in the μ direction. Similar to the radiation intensity, a function of wavelength, direction, and location can be written as follows [29]. The radiation energy balance terms are schematically presented in Figure 1.

$$(3) \quad J_\lambda(X,\mu,t) = +K_\lambda(\mu)I_{b,\lambda}(T(X,t)) + \frac{1}{2}\int_{\mu'=-1}^{\mu'=1}\sigma_{s\lambda}(\mu')\phi_\lambda(\mu \to \mu')I_\lambda(X,\mu',t)d\mu'$$

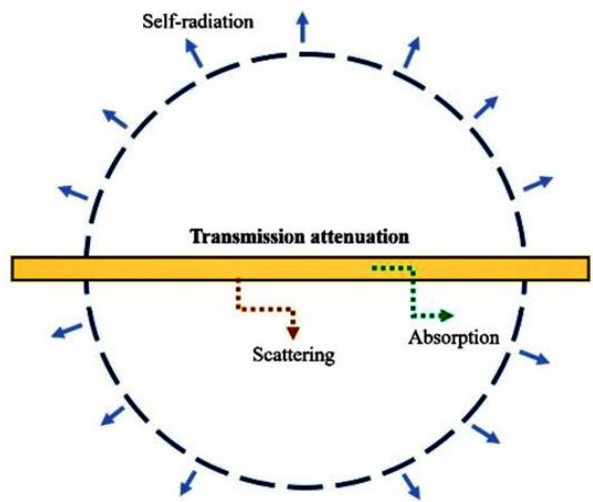

**Figure 1.** Radiative energy balance in a gas element.

The radiative boundary conditions expressed by integro-differential equations are as follows:

$$\begin{cases} I_\lambda(0,\mu,t) = I_{\lambda,b}(f(t)) \\ I_\lambda(E,\mu,t) = I_{\lambda,b}(T_E) \end{cases} \tag{4}$$

The monochromatic intensity is written by plank's law at the situation that the transparent of the black body, given as:

$$I(X,T(X,t)) = \frac{C_1}{\lambda^5 [\exp(\frac{C_2}{\lambda T(X,t)}) - 1]} \tag{5}$$

Where C1=1.19×10-16 ($\frac{W}{m^2}$), C2=1.44×10-2 (m.K). One dimensional combined transient energy equation is mentioned as follows [30].

$$\rho c_p \frac{\partial T(X,t)}{\partial t} = \nabla \cdot (K \nabla T(X,t) - \dot{Q}_r) + \dot{q}_{gen} \tag{6}$$

Where $\rho$ is density, cp specifies heat under constant pressure, T is temperature, t is time, K is thermal conductivity, $\dot{Q}_r$ shows gradient heat flux vector and the divergence of radiative heat flux in (6) is the net radiant energy emitted from a unit volume. It is a scalar quantity, a function of position, and is called a radiative source where Eq.7 present it. [30]

$$\dot{Q}_r(X) = 2\pi \int_{\lambda=0}^{\infty} \int_{\mu=-1}^{\mu=1} I(\lambda,X) \cdot \mu d\mu d\lambda \tag{7}$$

The unknown parameter of the radiation heat equation $I(\lambda,x)$ is the radiation intensity which is a function of direction µ at a point x. Radiation source in a one-dimensional heat transfer is given by Eq.8. [31]

$$\nabla \dot{Q}_r(X) = -\frac{d}{dX} \dot{Q}_r(X) \tag{8}$$

The following equations considered to describe radiative-conduction heat transfer through a thin fibrous media with homogeneous properties. The temperature distribution in the x-direction, T(x), is derived by solving the non-linear equation.

$$\rho c_p \frac{\partial T(X,t)}{\partial t} = \frac{\partial}{\partial X}(K(T(X,t)) \cdot \frac{\partial T(X,t)}{\partial t}) + \nabla \dot{Q}_r \tag{9}$$

The boundary conditions for solving the Eq.9 are written as;

$$T(0) = T_0 \tag{10}$$

Where E is the fibrous media length. The conduction heat transfer term defined as;

$$\dot{Q}_c(X,t) = -K(T(X,t)) \cdot \frac{\partial T(X,t)}{\partial X} \tag{11}$$

The thermal conductivity value ($\frac{w}{m \cdot K}$), is defined as a function of the absolute temperature T(K).

$$K(T) = K(298.15K) \sum_i (\frac{T}{298.15})^i \tag{12}$$

Also, the volumetric specific heat capacity values were defined as a function of absolute temperature T(K), which is given by Eq.13.

$$\rho c_p = (\rho c_p)(298.15K) \sum_i (\frac{T}{298.15})^t \tag{13}$$

In this problem, the combined heat flux is given by the summation of conduction and radiation. The unknown parameter after coupling Eq.7 and Eq.9 are monochromatic radiation intensity and temperature field, which are provided by solving Eq.14.

$$\rho c_p \frac{\partial T(X,t)}{\partial t} - \frac{\partial}{\partial X}(KT(X,t)\frac{\partial T(X,t)}{\partial t}) = S_r(X,t) \tag{14}$$

Boundary conditions for solving the non-linear Eq.14 are written as:

$$T(0,t) = f(t) = \begin{cases} 50t + 300 & 0 \text{ p } t \text{ p } 1 \\ 300 & t \text{ f } 1 \end{cases} \tag{15a}$$

$$T(E,t) = T_E \tag{15b}$$

$$T(X,0) = T_E \tag{15c}$$

For solving transient equations of the layer, assuming the layer is initial at uniform temperature TE and also a linear heat shock exists on the initial position of fibrous media. The temperature of the ends of the fibrous media is assumed constant.

*2.2 Deep Neural Network method*

The essential milestone in using an artificial neural network algorithm is developing the special applicable algorithm to find the best correlation between different essential parameters in a problem. By simulating a problem with discretizing the governing equations, the dataset was imported to the algorithm in order to find the best correlation and output data predictions.

The utilized AI algorithm for this research is Long Short-Term Memory (LSTM). It is one of the Deep Neural Network (DNN) methods which is so applicable for predicting the time series data. The data is collected during experiments or doing simulations are time-dependent. For this reason, LSTM is the most accurate algorithm for predicting output data of our recent research.

For storing the data, the "cell-states" is used to store the long-term data in hidden layers. As it is presented in Eq.16 and Eq.17, f_t and i_t presents the forget and input gates to control the input of each cell [33].

$$f_t = g(W_f \cdot [h_{t-1}.X_t] + b_f) \tag{16}$$

$$i_t = g(W_i \cdot [h_{t-1}.X_t] + b_i) \tag{17}$$

Also, $g$ is a non-linear sigmoid function utilized as an activation function here. Also, $W$ and b introduce the weight matrix and bias function, $h_{t-1}$ introduces the last time step output, and $X_t$ presents the current time step input.

Also, Eq.18 presents the relation to understand the input's current state,

$$\acute{C}_t = tanh(W_c \cdot [h_{t-1}.X_t] + b_c) \tag{18}$$

Where c index shows each parameter's current state. Eq.19 is considered as using both forget and input gates to obtain the current cell-state [33].

$$C_t = f_t * C_{t-1} + i_t * \acute{C}_t \tag{19}$$

By utilizing the output of each cell-state which is presented as Eq.20, the algorithm output is presented as Eq.21.

$$O_t = g(W_o \cdot [h_{t-1}.X_t] + b_o) \tag{20}$$

$$h_t = O_t * \tanh(C_t) \tag{21}$$

Also, o index introduces the cell-state output parameter.

The AI predictions are done using these equations of the LSTM algorithm. Furthermore, for the input's order reduction, the Mahalanobis distance method is used to reduce the prediction time [34,35]. Then, the simulation was compared with the AI-predicted data to find the highly accurate and fast one. For the recent study, the best learning rate was 0.01. The reason is that at this specific number, with low learning rates, the loss improves slowly, then training accelerates until the learning rate becomes too large and loss goes up so the training process diverges.

In this study, the temperature distribution of experimental measurements is predicted with DNN based algorithm. As it is shown in Figure 2, the utilized artificial structure consists of three main layers named as input layer, hidden layer, and output layer.

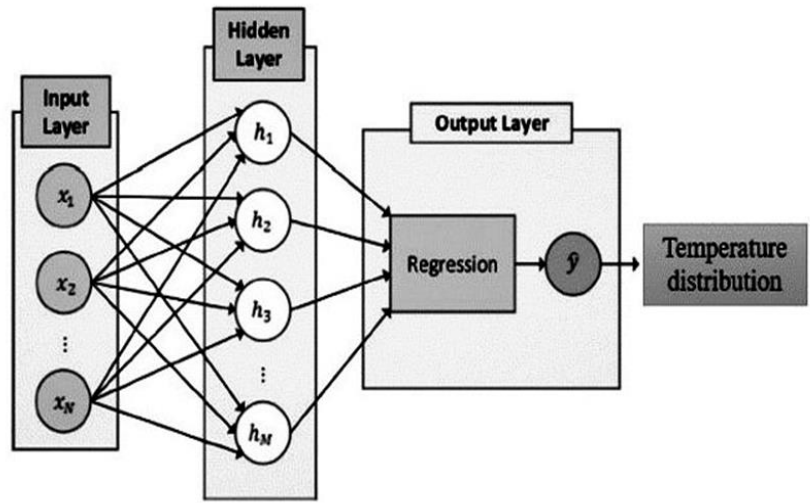

**Figure 2.** The DNN structure of recent study.

## 3. Results and discussion

This paper focused on one-dimensional combined heat transfer of PMMA-POF by computer numerical analysis and the LSTM method. The data is validated by Asllanaj et al. [36]. Figure 3 shows the temperature distribution of PMMA-POF during four predictions in different periods and compared with a previous experimental study [36]. Although the source temperature of the front layer of PMMA sample increases from 300K to 400K in a period of 1s, the temperature of the external wall remains constant at 300K.

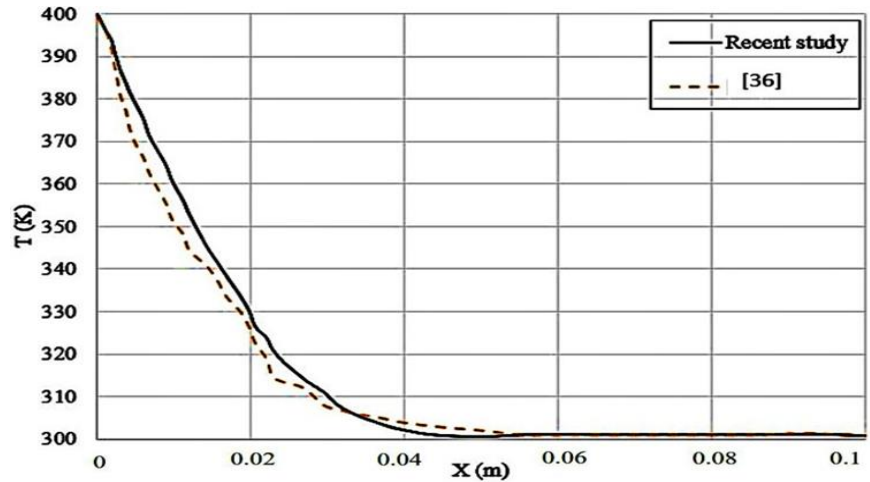

**Figure 3.** Comparing the results of transient thermal distribution in silica fibers.

As presented in Figure 3, the temperature gradient occurs from the beginning of the sample up to the 0.05 meter approximately. Then, the temperature remains constant up to the end (length=0.1 m). Moreover, the sample experienced a temperature decreasing trend, as same as Asllanaj [36] study. Although this study is verified with Asllanaj, two carves have some differences which are related to the research method. Two important points are at the beginning and end of the curves which are coincided.

Figure 4 presents the temperature distribution of PMMA-POF which is obtained from computer numerical analysis at various times. The utilized temperature values equal f(t) at x=0 and $T_E$ at x=1.

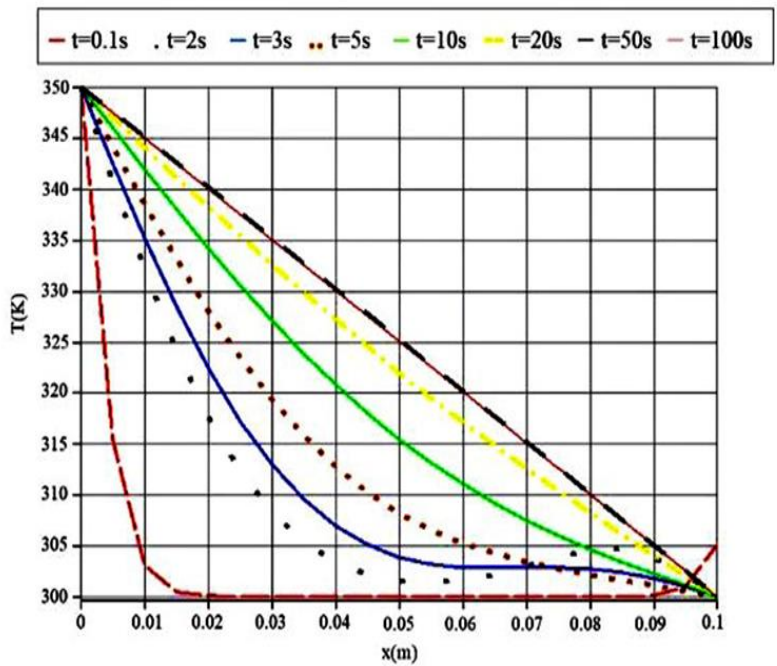

**Figure 4.** Temperature profile in various time steps.

As presented in Figure 4, for the maximum simulation time (t=100s), the decreasing trend of the temperature along the sample length is completely constant and the same as the steady condition. As the simulation time decrease, the temperature profile varies and in the minimum time, it experiences a sudden decrease in the first few lengths. The first and last points of simulation coincide, but as a result, curves vary with respect to the simulation time.

Figure 5 and Figure 6 show the heat fluxes variates with time and position. The conductive heat flux is similar to the radiative one in the case of behavior.

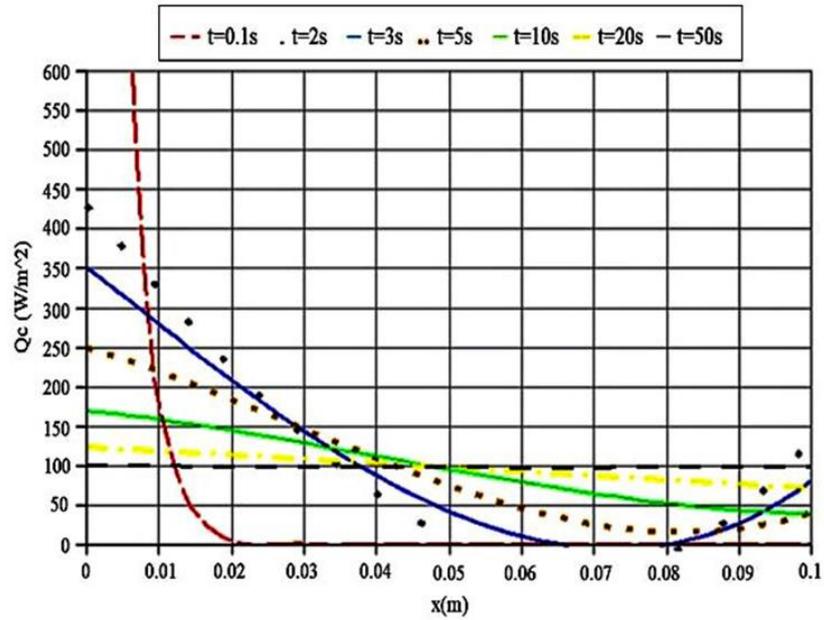

**Figure 5.** Conductive heat flux in various time steps.

As presented in Figure 5, the conductive heat flux converged to stable conditions around 100 ($\frac{W}{m^2}$) and it is stable approximately. This stable heat flux is obtained by the maximum simulation time. This figure presents the fact that for obtaining the stable and constant conductive heat transfer all along with the sample, the simulation time should be set at 50 s. By decreasing the simulation time, the output curves don't have a stable trend and their distances from the stable line increase.

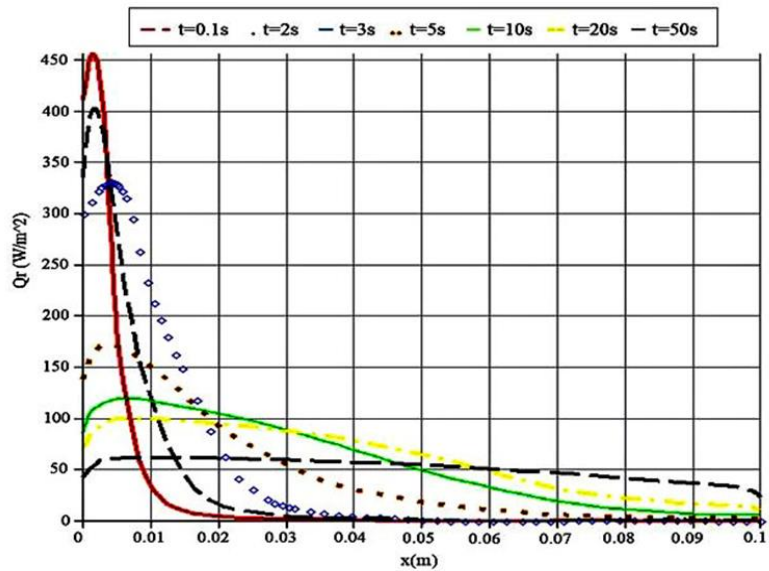

**Figure 6.** Radiative heat flux in various time steps.

As presented in Figure 6, the slight stable condition belongs to the maximum simulation time when the radiative heat transfer reaches 50 ($\frac{W}{m^2}$) approximately at 0.06m. Not only the radiative heat transfer can't be as same as conductive heat transfer, but also it has some slight decreasing trends. The basic rule which is related to the simulation time is as same as Figure 6 in the case of increasing its time, but when these two figures compare with each other, the amount of conductive heat transfer is considered twice the radiative heat transfer. As a matter of fact, in the same physical conditions, conduction plays a vital role in heat transfer from PMMA sample.

But when these solutions were done by an artificial intelligence method, the datasets have fewer than the expected data stores in the case of experimental data. Testing time (s), temperature distribution (K), radiative heat transfer ($\frac{W}{m^2}$), conductive heat transfer ($\frac{W}{m^2}$). The parameter's correlation expressed in a value between 0 to 1. To visualize a most correlated parameter by Pearson's correlation method, the correlation matrix should be prepared as Figure 7. This method utilized heat-map visualization in order to find the best relation between different parameters. It presents the values with different spectrums of dark and light colors. In this spectrum, the lighter one shows the best relation. Hence, it is obvious to prove the fact that by using the AI prediction method, we can find a good relationship between milestone parameters.

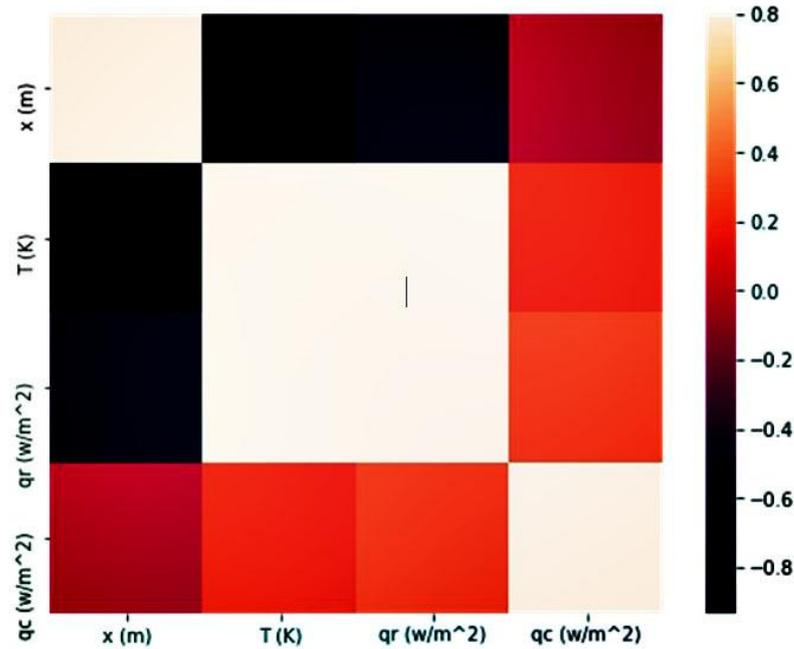

**Figure 7.** The correlation matrix between parameters.

As it can infer from Figure 7, the correlation matrix clarifies the relation of different parameters which are predicted from the LSTM method. This kind of presentation helps to better understand the intensity of these relations. The darkest color presents the least related one and the brightest color presents the most related one. The most relevant parameters are temperature and radiative heat transfer. Furthermore, the least relevant ones are the relation between sample length and temperature distribution, and the other one which is predicted by the LSTM method is the pair of radiative heat transfer and sample length.

Pearson's method of correlation is used in order to find the relation between all of the effective parameters of this study. Moreover, the LSTM predicted parameters are plotted as a matrix of figures. Hence, we can find that the numerical solution in this problem can measure positive magnitudes. Despite this, our utilized method predicts both negative and positive values simultaneously. Also, Figure 8 presents the linear regression between the correlation's scatter data. This figure is as same as Figure 9 but the data were best fitted.

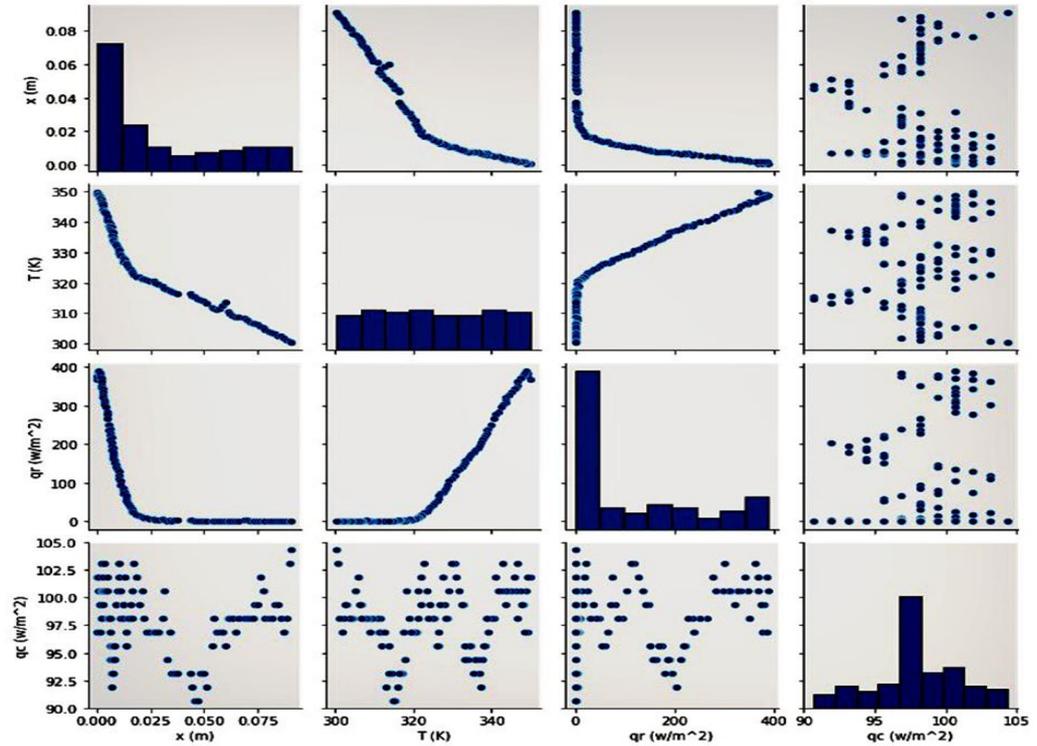

**Figure 8.** The scatter plot of different heat transfer parameters.

Figure 8 shows the predicted values as a correlation matrix of different variables at t=50s. The results show that by moving along the PMMA sample, the conduction and radiative heat transfer rates decrease and reach a constant value. Fixing these values leads to a decrease in temperature distribution. Each element of these rows and columns figure outs special relation between effective parameters. If it is imagined that the basic cross-line presents the relation of the same parameter. Going upward from the cross-line is presented the reverse relation of each downward element. For more details, the x was replaced with the y-axis in a reversing procedure.

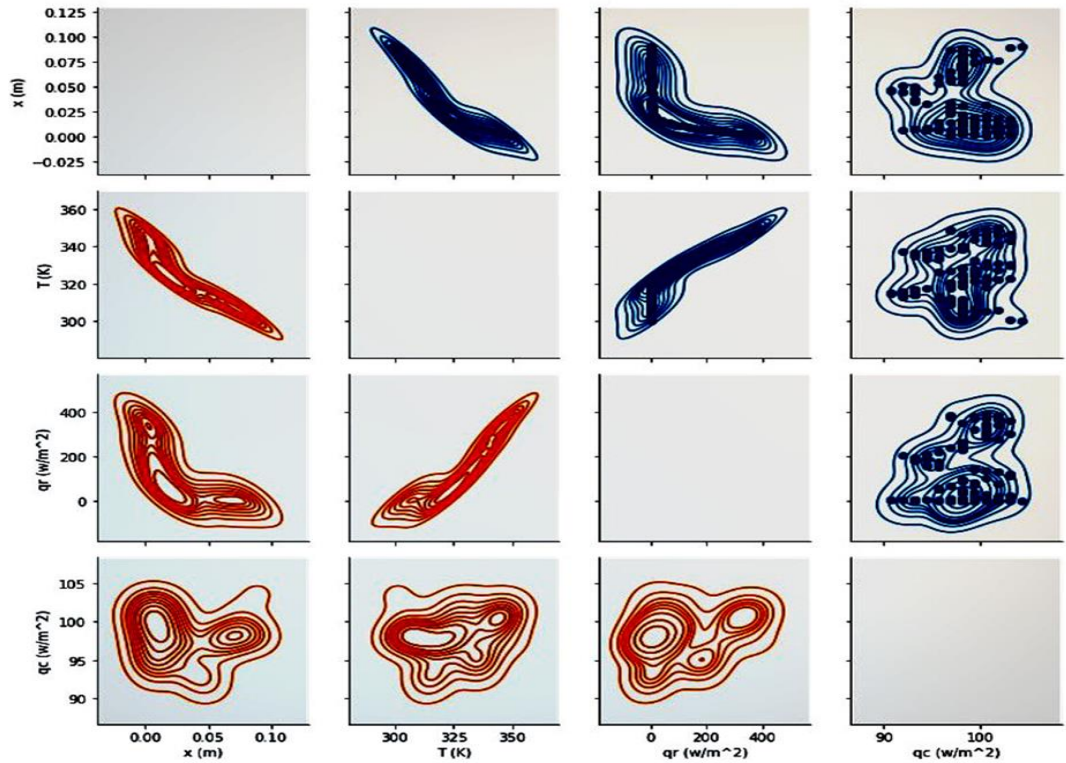

**Figure 9.** The overall figures of different parameter's correlations

Also, Figure 9 presents a gradient descent of different parameters by the cost function optimization. The gradient descent presents the possible answers. The compressed lines are the nearest answer between others. Imagine a hill, these compressed lines have the lowest value of gradient descent and scattered ones have the least amount of gradient descent. These charts specially Figure 9 present how the subset of scattering answers are optimized to find the best-fit ones.

The comparison between numerical modeling and experimental one is done and presented in Figure 3. As the machine learning method was utilized, it should be compared with numerical methods to find its approach in this study. Figure 10 presents this comparison well.

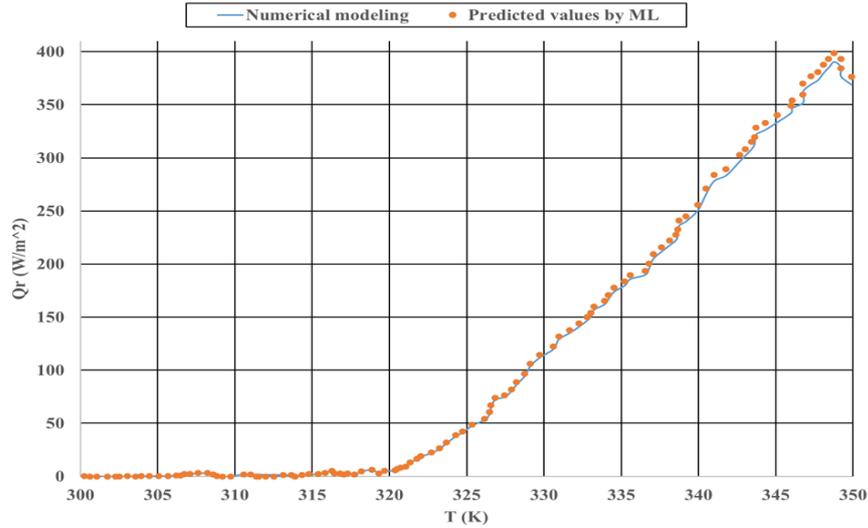

**Figure 10.** Comparison between two utilized different method.

As presented in Figure 10, predicted values coincide with the numerical method in most of the points. However, some tiny differences are observed; these are related to selecting a hyperparameter for the LSTM algorithm.

One of the vital tasks in machine learning implementation is to evaluate the utilized algorithm with ROC curve and confusion matrix (Figure 11). Furthermore, another one is to examine the accuracy of their performance. Utilized algorithms for estimating the conductive and radiative heat transfer were assessed, and the model performances are reported in Table 1. Statistical analysis was performed and assessed using MAE, root mean square error (RMSE), accuracy (ACC), specificity (FPR), sensitivity (TPR), positive predicted values (PPV), and true negative rate (TNR).

**Table 1.** The learning accuracy and its performance are evaluated by different measurement tools.

| Parameter | Method | TNR | PPV | TPR | FPR | ACC | RMSE | MAE |
| --- | --- | --- | --- | --- | --- | --- | --- | --- |
| $Q_c$ | LSTM | 0.98 | 0.98 | 0.94 | 0.02 | 0.96 | 16.42 | 0.06 |
| $Q_r$ | LSTM | 0.98 | 0.98 | 0.92 | 0.02 | 0.95 | 37.53 | 0.07 |

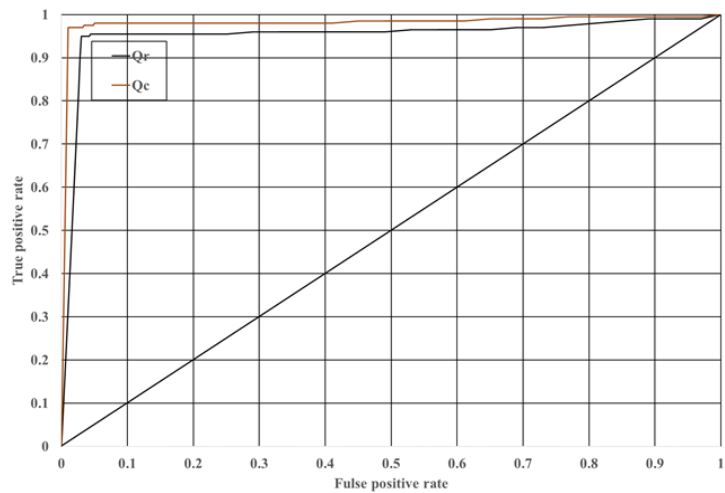

**Figure 11.** ROC curve and confusion matrix of radiative and conductive heat transfer prediction.

As presented in Figure 11 and Table 1, the confusion matrix can present whether LSTM algorithm is accurate enough to predict different heat transfer problem values. As it is demonstrated, the accuracy for both parameter predictions is 0.98. For the future research exploring further applications, e.g., [37-68] is proposed.

## 4. Conclusions

A new comparative artificial intelligence prediction method of the heat transfer modeling of PMMA samples based on coupled heat transfer is proposed and validated with the other work in this study. The analysis of combined heat transfer of PMMA sample as a core of fibrous media is solved by the transient coupled system of equations. It is done by computer implementation by using the implicit finite difference scheme for conduction and for the radiation heat flux by using the Gaussian numerical method. The temperature distribution is validated with perfect agreement. The results present an intense gradient of the temperature values in the primary positions, which makes the spectral amount of conductive heat flux and the heat flux of two heat transfer types. Hence, it depends on temperature distribution becoming steady states at the same time.

It is proved that the conductive and radiative heat flux is similar in the case of trending behavior, but it is twice an amount approximately. Furthermore, their total flux becomes approximately constant at the steady-state condition. The artificial intelligence method and numerical results demonstrate good agreement and verify the novel LSTM algorithm in future research. The overall calculation time is increased to 2s by using this method. As compared in this study, the LSTM method can easily predict and evaluate these kinds of problems in different conditions, and also, it is the most accurate one (98%), which is evaluated by confusion matrix and ROC curve. This method can predict the temperature distribution in a PMMA sample. It will help to find the best industrial results in less time. It improves the prediction speed when compared with numerical solutions.

**Author Contributions:** Authors make equal contribution.

**Funding:** This research had been funded by H2020.

**References**

1.	Yang D, Yu J, Tao X, et al. Structural and mechanical properties of polymeric optical fiber. Materials Science and Engineering: A. 2004;364(1-2):256-259.
2.	Suchorab Z, Franus M, Barnat-Hunek D. Properties of Fibrous Concrete Made with Plastic Optical Fibers from E-Waste. Materials. 2020;13(10):2414.
3.	Yuan H, Wang Y, Zhao R, et al. An anti-noise composite optical fiber vibration sensing System. Optics and Lasers in Engineering. 2021;139:106483.
4.	Chaitanya S, Mukherjee G, Banerjee M, et al. Optical studies of Rhodamine B doped polymethyl methacrylate (PMMA) films. Materials Today: Proceedings. 2020.
5.	Bzówka J, Grygierek M, Rokitowski P. Experimental investigation using distributed optical fiber sensor measurements in unbound granular layers. Engineering Structures. 2021;231:111767.
6.	Jderu A, Soto MA, Enachescu M, et al. Liquid Flow Meter by Fiber-Optic Sensing of Heat Propagation. Sensors. 2021;21(2):355.
7.	AlAbdulaal T, Yahia I. Optical linearity and nonlinearity, structural morphology of TiO2-doped PMMA/FTO polymeric nanocomposite films: Laser power attenuation. Optik. 2021;227:166036.
8.	Jatoi AS, Khan FSA, Mazari SA, et al. Current applications of smart nanotextiles and future trends. Nanosensors and Nanodevices for Smart Multifunctional Textiles: Elsevier; 2021. p. 343-365.
9.	Cárdenas D, Gaudino R, Nespola A, et al., editors. 10Mb/s Ethernet transmission over 425m of large core Step Index POF: a media converter prototype. International Conference on Plastic Optical Fiber (ICPOF 2006); 2006.
10.	Hemath M, Mavinkere Rangappa S, Kushvaha V, et al. A comprehensive review on mechanical, electromagnetic radiation shielding, and thermal conductivity of fibers/inorganic fillers reinforced hybrid polymer composites. Polymer Composites. 2020;41(10):3940-3965.
11.	Sallam O, Madbouly A, Elalaily N, et al. Physical properties and radiation shielding parameters of bismuth borate glasses doped transition metals. Journal of Alloys and Compounds. 2020:156056.
12.	Li D, Zhang C, Li Q, et al. Thermal performance evaluation of glass window combining silica aerogels and phase change materials for cold climate of China. Applied Thermal Engineering. 2020;165:114547.
13.	Barnoss S, Aribou N, Nioua Y, et al. Dielectric Properties of PMMA/PPy Composite Materials. Nanoscience and Nanotechnology in Security and Protection against CBRN Threats: Springer; 2020. p. 259-271.

14. Sans M, Schick V, Parent G, et al. Experimental characterization of the coupled conductive and radiative heat transfer in ceramic foams with a flash method at high temperature. International Journal of Heat and Mass Transfer. 2020;148:119077.
15. Malek M, Izem N, Mohamed MS, et al. Numerical solution of Rosseland model for transient thermal radiation in non-grey optically thick media using enriched basis functions. Mathematics and Computers in Simulation. 2021;180:258-275.
16. Satapathy AK, Nashine P. Solving Transient Conduction and Radiation Using Finite Volume Method. International Journal of Mechanical and Mechatronics Engineering. 2015;8(3):645-649.
17. Wakif A. A novel numerical procedure for simulating steady MHD convective flows of radiative Casson fluids over a horizontal stretching sheet with irregular geometry under the combined influence of temperature-dependent viscosity and thermal conductivity. Mathematical Problems in Engineering. 2020;2020.
18. Makinde O. Free convection flow with thermal radiation and mass transfer past a moving vertical porous plate. International Communications in Heat and Mass Transfer. 2005;32(10):1411-1419.
19. Chu Y-M, Nazeer M, Khan MI, et al. Combined impacts of heat source/sink, radiative heat flux, temperature dependent thermal conductivity on forced convective Rabinowitsch fluid. International Communications in Heat and Mass Transfer. 2020:105011.
20. HERNANDEZ D, Denis Y. Energy Management System Industrialization for Off-Grids Power Systems Based on Data-Driven Machine Learning Models.
21. Kwon B, Ejaz F, Hwang LK. Machine learning for heat transfer correlations. International Communications in Heat and Mass Transfer. 2020;116:104694.
22. Potočnik P, Škerl P, Govekar E. Machine-learning-based multi-step heat demand forecasting in a district heating system. Energy and Buildings. 2021;233:110673.
23. Zhao J, Ye F. Where ThermoMesh meets ThermoNet: A machine learning based sensor for heat source localization and peak temperature estimation. Sensors and Actuators A: Physical. 2019;292:30-38.
24. Ju S, Shiomi J. Materials informatics for heat transfer: Recent progress and perspectives. Nanoscale and Microscale Thermophysical Engineering. 2019 Apr 3;23(2):157-72.
25. Ju S, Shimizu S, Shiomi J. Designing thermal functional materials by coupling thermal transport calculations and machine learning. Journal of Applied Physics. 2020 Oct 28;128(16):161102.
26. Acakpovi A, Matoumona PLMV, editors. Comparative analysis of plastic optical fiber and glass optical fiber for home networks. 2012 IEEE 4th International Conference on Adaptive Science & Technology (ICAST); 2012: IEEE.
27. Siegel R. Thermal radiation heat transfer. CRC press; 2001.
28. Lockhat R. Physics: Wheatstone bridge. Southern African Journal of Anaesthesia and Analgesia. 2020;26(6 Suppl 3):S100-101.
29. Modest MF. Radiative heat transfer. Academic press; 2013.
30. Ozisik MN. Radiative transfer and interactions with conduction and convection (Book- Radiative transfer and interactions with conduction and convection.). New York, Wiley-Interscience, 1973 587 p. 1973.
31. Kant K, Shukla A, Sharma A, et al. Heat transfer studies of photovoltaic panel coupled with phase change material. Solar Energy. 2016;140:151-161.
32. Assael M, Botsios S, Gialou K, et al. Thermal conductivity of polymethyl methacrylate (PMMA) and borosilicate crown glass BK7. International Journal of Thermophysics. 2005;26(5):1595-1605.
33. Aslam M, Lee J-M, Kim H-S, et al. Deep learning models for long-term solar radiation forecasting considering microgrid installation: A comparative study. Energies. 2020;13(1):147.
34. Duan Z, Yang Y, Zhang K, et al. Improved deep hybrid networks for urban traffic flow prediction using trajectory data. Ieee Access. 2018;6:31820-31827.
35. Zucatti V, Lui HF, Pitz DB, et al. Assessment of reduced-order modeling strategies for convective heat transfer. Numerical Heat Transfer, Part A: Applications. 2020;77(7):702-729.
36. Asllanaj F, Jeandel G, Roche JR, et al. Transient combined radiation and conduction heat transfer in fibrous media with temperature and flux boundary conditions, International Journal of Thermal Sciences; 2004; 43: 939–950.
37. Samadianfard, Saeed, et al. "Wind speed prediction using a hybrid model of the multi-layer perceptron and whale optimization algorithm." Energy Reports 6 (2020): 1147-1159.
38. Taherei Ghazvinei, Pezhman, et al. "Sugarcane growth prediction based on meteorological parameters using extreme learning machine and artificial neural network." Engineering Applications of Computational Fluid Mechanics 12.1 (2018): 738-749.
39. Kargar, Katayoun, et al. "Estimating longitudinal dispersion coefficient in natural streams using empirical models and machine learning algorithms." Engineering Applications of Computational Fluid Mechanics 14.1 (2020): 311-322.
40. Qasem, Sultan Noman, et al. "Estimating daily dew point temperature using machine learning algorithms." Water 11.3 (2019): 582.
41. Mosavi, Amir, and Atieh Vaezipour. "Reactive search optimization; application to multiobjective optimization problems." Applied Mathematics 3.10A (2012): 1572-1582.
42. Shabani, Sevda, et al. "Modeling pan evaporation using Gaussian process regression K-nearest neighbors random forest and support vector machines; comparative analysis." Atmosphere 11.1 (2020): 66.

43. Ghalandari, Mohammad, et al. "Aeromechanical optimization of first row compressor test stand blades using a hybrid machine learning model of genetic algorithm, artificial neural networks and design of experiments." Engineering Applications of Computational Fluid Mechanics 13.1 (2019): 892-904.
44. Mosavi, Amir. "Multiple criteria decision-making preprocessing using data mining tools." arXiv preprint arXiv:1004.3258 (2010).
45. Mahmoudi, Mohammad Reza, et al. "Principal component analysis to study the relations between the spread rates of COVID-19 in high risks countries." Alexandria Engineering Journal 60.1 (2021): 457-464.
46. Karballaeezadeh, Nader, et al. "Prediction of remaining service life of pavement using an optimized support vector machine (case study of Semnan–Firuzkuh road)." Engineering Applications of Computational Fluid Mechanics 13.1 (2019): 188-198.
47. Moeini, Iman, et al. "Modeling the time-dependent characteristics of perovskite solar cells." Solar Energy 170 (2018): 969-973.
48. Shamshirband, Shahaboddin, et al. "Prediction of significant wave height; comparison between nested grid numerical model, and machine learning models of artificial neural networks, extreme learning and support vector machines." Engineering Applications of Computational Fluid Mechanics 14.1 (2020): 805-817.
49. Samadianfard, Saeed, et al. "Support vector regression integrated with fruit fly optimization algorithm for river flow forecasting in Lake Urmia Basin." Water 11.9 (2019): 1934.
50. Lei, Xinxiang, et al. "GIS-based machine learning algorithms for gully erosion susceptibility mapping in a semi-arid region of Iran." Remote Sensing 12.15 (2020): 2478.
51. Adejuwon, Adeyemi, and Amir Mosavi. "Domain driven data mining: application to business." International Journal of Computer Science Issues 7.4 (2010): 41-44.
52. Riahi-Madvar, Hossien, et al. "Comparative analysis of soft computing techniques RBF, MLP, and ANFIS with MLR and MNLR for predicting grade-control scour hole geometry." Engineering Applications of Computational Fluid Mechanics 13.1 (2019): 529-550.
53. Mosavi, Amir, and Mohammad Edalatifar. "A hybrid neuro-fuzzy algorithm for prediction of reference evapotranspiration." International conference on global research and education. Springer, Cham, 2018.
54. Asadi, Esmaeil, et al. "Groundwater quality assessment for sustainable drinking and irrigation." Sustainability 12.1 (2019): 177.
Mosavi, Amir, and Abdullah Bahmani. "Energy consumption prediction using machine learning; a review." (2019).
55. Dineva, Adrienn, et al. "Review of soft computing models in design and control of rotating electrical machines." Energies 12.6 (2019): 1049.
56. Mosavi, Amir, and Timon Rabczuk. "Learning and intelligent optimization for material design innovation." In International Conference on Learning and Intelligent Optimization, pp. 358-363. Springer, Cham, 2017.
57. Torabi, Mehrnoosh, et al. "A hybrid machine learning approach for daily prediction of solar radiation." International Conference on Global Research and Education. Springer, Cham, 2018.
58. Mosavi, Amirhosein, et al. "Comprehensive review of deep reinforcement learning methods and applications in economics." Mathematics 8.10 (2020): 1640.
59. Ahmadi, Mohammad Hossein, et al. "Evaluation of electrical efficiency of photovoltaic thermal solar collector." Engineering Applications of Computational Fluid Mechanics 14.1 (2020): 545-565.
60. Ghalandari, Mohammad, et al. "Flutter speed estimation using presented differential quadrature method formulation." Engineering Applications of Computational Fluid Mechanics 13.1 (2019): 804-810.
61. Ijadi Maghsoodi, Abteen, et al. "Renewable energy technology selection problem using integrated h-swara-multimoora approach." Sustainability 10.12 (2018): 4481.
62. Mohammadzadeh S, Danial, et al. "Prediction of compression index of fine-grained soils using a gene expression programming model." Infrastructures 4.2 (2019): 26.
63. Sadeghzadeh, Milad, et al. "Prediction of thermo-physical properties of TiO2-Al2O3/water nanoparticles by using artificial neural network." Nanomaterials 10.4 (2020): 697.
64. Choubin, Bahram, et al. "Earth fissure hazard prediction using machine learning models." Environmental research 179 (2019): 108770.
65. Emadi, Mostafa, et al. "Predicting and mapping of soil organic carbon using machine learning algorithms in Northern Iran." Remote Sensing 12.14 (2020): 2234.
66. Shamshirband, Shahaboddin, et al. "Developing an ANFIS-PSO model to predict mercury emissions in combustion flue gases." Mathematics 7.10 (2019): 965.
67. Salcedo-Sanz, Sancho, et al. "Machine learning information fusion in Earth observation: A comprehensive review of methods, applications and data sources." Information Fusion 63 (2020): 256-272.
78. Mousavi, Seyed Milad, et al. "Deep learning for wave energy converter modeling using long short-term memory." Mathematics 9.8 (2021): 871.